# Hierarchical Bayesian Framework for Multisource Domain Adaptation

Alexander M Glandon
*Electrical and Computer Engineering Dept.*
Old Dominion University
Norfolk, U.S.
aglan001@odu.edu

Khan M Iftekharuddin
*Electrical and Computer Engineering Dept.*
Old Dominion University
Norfolk, U.S.
kiftekha@odu.edu

*Abstract*—Multisource domain adaptation (MDA) aims to use multiple source datasets with available labels to infer labels on a target dataset without available labels for target supervision. Prior works on MDA in the literature is ad-hoc as the pretraining of source models is either based on weight sharing or uses independently trained models. This work proposes a Bayesian framework for pretraining in MDA by considering that the distributions of different source domains are typically similar. The Hierarchical Bayesian Framework uses similarity between the different source data distributions to optimize the pretraining for MDA. Experiments using the proposed Bayesian framework for MDA show that our framework improves accuracy on recognition tasks for a large benchmark dataset. Performance comparison with state-of-the-art MDA methods on the challenging problem of human action recognition in multi-domain benchmark Daily-DA RGB video shows the proposed Bayesian Framework offers a 17.29% improvement in accuracy when compared to the state-of-the-art methods in the literature.

## I. Introduction

Deep learning algorithms have been used extensively in recent years for learning from data with few or no labels, known as semi-supervised or unsupervised learning. Early research with a single source dataset and a single target dataset involves unsupervised domain adaptation (UDA) for learning. Reference [1] presents discriminator-based approaches to align the source and the target features while [2] shows the discriminator approach for video action recognition. Aligning Correlation Information [3] considers spatiotemporal pixels to align the source and the target domains to potentially find correlation between different frames. In Open Compound UDA [4], a related problem is solved where the target data is unlabeled, and the target data comes from domains that are not specified and possibly heterogeneous.

Since there are typically multiple datasets for a given problem (with different distributions), it is natural to use multiple sources to learn about an unlabeled target dataset. This problem is known in the literature as multisource domain adaptation (MDA). Existing MDA approaches pretrain feature extractors on all source domains and perform some domain alignment. In the Deep Cocktail Network [9] approach, the first step is to jointly train a feature extractor and class discriminator using the source images. This step is followed by multi-way adversarial adaptation which trains discriminators to detect the similarity of samples to each source domain. A Deep Ensemble Learning method [10] trains encoders for each source domain and aligns the extracted features by enforcing a reconstruction of the samples using a decoder for each pair of source domains. Distilling Domain Adaptation [11] uses a discriminator to measure similarity between a source and a target sample, and focuses on training source classifiers with samples that are similar to target samples, thereby teaching the classifiers to understand samples that are similar to the target samples.

Another collection of works uses moment matching to align features. In Temporal Attentive Moment Alignment Network [12] instead of aligning the moments of entire videos, the authors align the moments of both spatial and temporal features. Source-Free Domain Adaptation [13] extends the work in [12] to scenarios where the source data is not available after the source model in trained (only for the UDA scenario). In another attention model, Attention-Based MDA [14], moment alignment is again combined with attention based learning. In Partial Feature Selection [15], features are selected and aligned using source and target domain moments. In Moment Matching for MDA [16], again using moments, each pair of source domains and source and target domains are aligned.

Bayesian methods have also been applied to MDA. Bayesian MDA [17] trains classifiers on each source domain and uses the similarity between the samples from the source and target to weigh each source classifier, thereby establishing a prior over the sources. Bayesian Uncertainty Matching [18] notes that the prediction uncertainty can be appended to encoder features for source and target samples and used to train a discriminator for domain alignment. Bayesian Domain Adaptation with Gaussian Mixture [19] considers the idea of using a Bayesian model to capture the multiple source domains. However, these works do not use deep neural networks within their Bayesian framework to solve more complex datasets with rich information. Overall, the existing methods either use a single feature extractor for all source domains or use different feature extractors for each source domain.

Consequently, this work proposes a Hierarchical Bayesian Framework for learning a set of pretrained classifiers that can be used in various MDA schemes. Our proposed framework is shown to leverage the Bayesian framework in the context of training a deep neural network for the difficult task of video perception. Our method is based on finding a balance between training a single source encoder [9] [12] [14] [16] on one extreme and training $J$ source encoders for $J$ source domains [10] [11] [15] on the other extreme. Hierarchical Bayesian modeling allows us to tune this balance. Methods that use a pretrained source encoder as the first part of the pipeline can be improved if a stronger pretrained encoder is developed using Hierarchical Bayesian. This can improve the result of MDA not only for action videos [12], but for any data domain where a pretrained feature extractor is used (e.g. digits MDA in [9] [11] [14] [15] and other types of image MDA in [9] [10] [14] [15] [16] [18]).

Our contributions include first a mathematical Bayesian formulation for the MDA problem. Second, we demonstrate that our Hierarchical Bayesian Model pretrained encoder improves the final MDA performance by comparing the result to the naïve cases of a single encoder and $J$ independently trained encoders per $J$ source domains (using the same architecture and same dimension of the parameters). This result applies to a large class of models, where the edge cases of our framework represent the existing results in literature. Third, we propose a confidence-based method to leverage the Hierarchical Bayesian Model in target domain inference. Fourth, to demonstrate the effectiveness, we apply our framework to a challenging problem of human action recognition in RGB video. We compare our results on a standard benchmark multisource human action dataset called Daily-DA to the results of the existing MDA methods and show a 17.29% improvement in action recognition accuracy.

Section II gives a background of existing methods that are fundamental to our improved approach to MDA video action recognition. Section III gives the mathematical hierarchical model description, derivation of the cost function from the model, and confidence-based target domain classification. We describe how we use an attention-based model as our deep model architecture. We also describe the benchmark human action recognition data that we used to test our Hierarchical Bayesian Framework. Section IV details the experiment and ablation study that is performed. We then present and discuss the results by comparing our performance to existing work. Section V concludes and suggests future directions.

## II. BACKGROUND

Our approach is developed based on Hierarchical Bayesian multitask learning. We review this model briefly and then give some example models from the literature for the MDA problem. In the method section we will show how these two approaches are combined.

### A. Hierarchical Bayesian Multitask Learning

Hierarchical Bayesian Multitask Learning, is originally proposed in [20] to solve the multi-task learning problem, which in our formulation consists of finding a model $G: \mathcal{X} \times \mathcal{J} \mapsto \mathcal{L}$ which takes a video $x \in \mathcal{X}$ and a domain index $j \in \mathcal{J}$, and predicts a label $y \in \mathcal{L}$. Multitask learning is different in that it provides a model that can predict labels for multiple source domains, as opposed to predicting labels for a single target domain. Authors sometimes refer to multitask learning as domain adaptation, such as above in [20] or in [21] where source and target are labeled and the goal is target domain performance. In our work we differentiate the domain adaptation problem to mean that target domain labels are not available.

The particular form of multitask learning in [20] is formulated using a Hierarchical Bayesian Model (HBM) given as follows. Given classifiers $F_1, \ldots, F_J$ for domains $D_1, \ldots, D_J$ let the classifier parameters be denoted $\theta_1, \ldots, \theta_J$. For each classifier, allow a prior distribution on the parameter vector to be multivariate normal as $\theta_j \sim MVN(\theta_*, \sigma_j^2 I)$. Note that each prior distribution $\theta_j$ has the same mean $\theta_*$. In practice, the value of $\sigma_j$ will be set to the same $\sigma_*$ for all domains $j$. Finally, $\theta_*$ is given a second order prior distribution as multivariate normal $\theta_* \sim MVN(0, \sigma_{**}^2 I)$ to regularize the entire network toward smaller parameter values. The values of $\sigma_*$ and $\sigma_{**}$ which control the spread of the multivariate normal distributions are assigned as hyperparameters in the optimization. A smaller value of $\sigma_*$ forces the models $F_{j_1}$, $F_{j_2}$ for each pair of domains to have more similar parameters $\theta_{j_1}, \theta_{j_2}$. A smaller value for $\sigma_{**}$ regularizes all of the models $F_j$ which reduces the norm of all parameter vectors $\theta_j$. The overall model G as defined above is given in eq. 1. after training using the corresponding loss function.

$$G(x,j) = F_j(x|\theta_j). \qquad (1)$$

Following [20], the loss function for the hierarchical multitask model is given in eq. 2. where the outer sum is taken over the $J$ domains indexed by $j$ and the inner sum is taken over the elements of the parameter vectors indexed by $n$.

$$loss(D_1, \ldots, D_J, \theta_1, \ldots \theta_J) = \qquad (2)$$

$$\sum_j \left( \mathbb{E}_{X,Y \sim D_j}[loss(F_j(X|\theta_j), Y)] + \sum_n \frac{(\theta_{jn} - \theta_{*n})^2}{2\sigma_j^2} \right) + \sum_n \frac{(\theta_{*n})^2}{2\sigma_{**}^2}.$$

We describe in the Methods section how we use the center model $\theta_*$ and individual models $\theta_1, \ldots, \theta_J$ to train classifiers in the MDA setting.

### B. Multisource Domain Adaptation

Existing MDA approaches are based on either pretraining a single source encoder [9] [12] [14] [16] or pretraining $J$ source encoders for $J$ source domains [10] [11] [15]. Here we review in detail an example of a model with a single encoder, Deep Cocktail Network. Then we review an example of a model with $J$ source encoders, Deep Ensemble Learning.

*1) Deep Cocktail Network*

Deep Cocktail Network [9] is an discriminator based approach to MDA. There is a single encoder model $F_E$ and $J$ decoding classifiers $F_{C1}, \ldots, F_{CJ}$ for each source domain. The model is trained together on all source domains with source samples from domain $D_j$ passing through encoder $F_E$ to extract features and then through respective decoder $F_{Cj}$ to perform classification in pretraining. Next, discriminator training is performed using source data and unlabeled target data to learn models $H_1, \ldots H_J$. Discriminator $H_j$ is trained to determine if sample $x$ is from domain $j$ or from the target domain $t$. Finally, in testing, sample $x_t$ is passed through encoder $F_E$ and in parallel through each decoder $F_{C1}, \ldots, F_{CJ}$ to give $J$ predictions. The predictions are combined in a weighted average with weights determined by the predicted confidence from each discriminator $H_1, \ldots H_J$ that the sample $x_t$ comes from domains $D_1, \ldots, D_J$ respectively. This confidence reflects the sample's similarity to samples from each source domain, and therefore the domains with higher similarity are given more weight to their prediction from $F_{Cj}$. In our methods we will use a similar confidence-based selection of source domain predictions. The important point for this method is that there is a single encoder for all $J$ source domains.

*2) Deep Ensemble Learning*

Deep Ensemble Learning [10] is another discriminator based approach to MDA. Here there are $J$ encoders $F_{E1}, \ldots, F_{EJ}$ for each source domain. In training, samples from respective sources are fed through the encoders to obtain source features. A loss term penalizes source features' differences to align the features extracted from all source domains. A set of $J(J-1)$ decoders $F_{Dj_1j_2}$ is used to reconstruct the inputs from domain $j_1$ to cross domain $j_2 \neq j_1$. Ensuring that the features from domain $j_1$ can be used to reconstruct inputs from domain $j_2$ gives addition alignment between domains. There are $J$ classifiers $F_{C1}, \ldots, F_{CJ}$ trained for each source domain $j$ given the features from source encoder $F_{Ej}$ and corresponding labels. Finally, given the target domain unlabeled inputs, a loss is used that penalizes classifier difference between $F_{Cj_1}$ and $F_{Cj_2}$, to align the classifier predictions on the target domain $t$. In testing, the classifier with maximum confidence is used to make the prediction. As we mentioned above, this technique will be used in our testing as well. The important point for this method is that there are $J$ independently trained source encoders for the $J$ source domains.

*3) Moment Alignment Network*

Similarly to Deep Cocktail Network [9], the Moment Alignment Network [12] implements a single shared encoder representation to extract spatial features from action videos. Then spatial and temporal moments are aligned to encourage alignment amount multiple source domains. Again, the classifier with higher certainty is given more weight.

*C. Daily-DA Benchmark Action Recognition Dataset*

Existing work on MDA has considered category analysis from text [20], sentiment analysis from text [17], digit classification from images [9] [11] [14] [15], object classification from images [9] [10] [14] [15] [16] [18], and human action recognition from videos [12]. In this work we focus on the challenging problem of human action recognition.

We use the Daily-DA benchmark action recognition dataset [12]. Several groups [13] [22] [23] [24] have used Daily-DA as a UDA task by training using a single source domain and testing on another target domain. Other work [25] [26] has been done on few shot learning, again for single source domain. A related problem is continuous video adaptation which is an online learning approach which was also tested on Daily-DA [27]. Finally, existing work uses Daily-DA for MDA in [12] for human action recognition. This is the method that we compare our result against in section IV.

III. METHODS

The Hierarchical Bayesian model is a general framework that can be integrated into different MDA schemes. First, we develop the general framework, and then we describe how we use this framework with a state-of-the-art deep attention model for video action recognition.

*A. Hierarchical Bayesian Multisource Domain Adaptation*

We adapt the Hierarchical Bayesian multitask learning [20] from section IIA for use in MDA. We start with the definition of multisource domain adaptation from section I. The input space is $\mathcal{X}$, the labels set is $\mathcal{L} = \{l_1, \ldots, l_K\}$, and there are $J$ source domains $D_1, \ldots, D_J$. Samples from domain $j$ are of the form $(x_{ji}, y_{ji}) \in D_j$, where $x_{ji}$ are examples from $\mathcal{X}$ with labels $y_{ji}$ from $\mathcal{L}$. There is a target domain with samples of the form $(x_{ti}, y_{ti}) \in D_t$. Using the source samples and labels and using only the target samples, we predict labels of the target samples with a model F: $\mathcal{X} \mapsto \mathcal{L}$. Generally, MDA methods are based on encoder and decoder models for feature extraction and classification respectively. Therefore, each model F can be written as the composition of an encoder and a decoder in eq. 3. and represents classification of the features obtained from the encoder.

$$F(x|\theta) = F_C(F_E(x|\theta_E)|\theta_C). \quad (3)$$

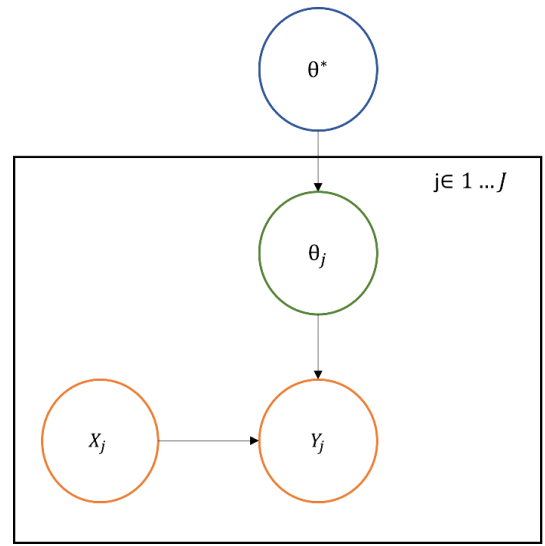

Fig. 1. Bayesian Network representation of Hierarchical Bayesian Model for Multisource Domain Adaptation

We present here the general framework for Hierarchical Bayesian MDA where an encoder and decoder are used, so that existing methods can be improved by using our framework. In section IIIB below we use the Bayesian model with an end-to-end network, without breaking the model into an encoder and decoder. If it is appropriate to the method being modified using our Bayesian approach, alignment of features based on feature moments, feature or class difference penalties, or domain discriminators can be obtained using additional loss terms like in [9] [10] [11] [12] [14] [15] [16] [18]. We use a set of classifiers $F_1, \ldots, F_J$ for domains $D_1, \ldots, D_J$ that are composed of encoders $F_{E1}, \ldots, F_{EJ}$ and decoders $F_{C1}, \ldots, F_{CJ}$. Encoder $F_{Ej}$ has parameters $\theta_{Ej}$ and decoder $F_{Cj}$ has parameters $\theta_{Cj}$. The component models can be any function where the gradient can be computed. We will use a convolutional and attention based deep neural network described in section IIID below. The output of our classifier is a probability distribution over possible class labels which we model as a categorical distribution $\hat{y} = F_C(F_E(x|\theta_E)|\theta_C)$. This means our predictions for each class are normalized to sum to 1, so that the prediction $\hat{y}_k = P(y_k = l_k)$ is a proper categorical distribution given by our prediction output vector. In the setting of $J$ source domains, the prediction for sample $x_{ji}$ is the categorical distribution over labels given by $\hat{y}_{kji}$ in eq. 4.

$$\hat{y}_{kji} = F(x_{ji}|\theta_j) = F_{Cj}\big(F_{Ej}(x_{ji}|\theta_{Ej})|\theta_{Cj}\big). \quad (4)$$

The Hierarchical Bayesian model can be represented visually as a Bayesian network [28], where a variable that depends on another variable has an arrow pointing to the dependent variable. This means the overall model can be factored into a product of conditional distributions as shown in Fig. 1 above. The prediction $\hat{y}_{kji}$ is given by the product of conditional distributions in eq. 5. below.

$$\hat{y}_{kji} = P(y_{kji} = l_k|x_{ji}, \theta_j)P(\theta_j|\theta_*)P(\theta_*). \quad (5)$$

Subsequently, the probability that sample $x_{ji}$ has label $l_k$, written in eq. 5. above $P(y_{kji} = l_k|x_{ji}, \theta_j)$ will be abbreviated

$$P(\theta_1 \ldots \theta_J, \theta^*|y_{11} \ldots y_{1I_1}, x_{11} \ldots x_{1I_1}, \ldots, y_{J1} \ldots y_{JI_J}, x_{J1} \ldots x_{JI_J}) \quad (6)$$

$$= \frac{P(y_{11} \ldots y_{1I_1}, x_{11} \ldots x_{1I_1}, \ldots, y_{J1} \ldots y_{JI_J}, x_{J1} \ldots x_{JI_J}, \theta_1 \ldots \theta_J, \theta^*)}{P(y_{11} \ldots y_{1I_1}, x_{11} \ldots x_{1I_1}, \ldots, y_{J1} \ldots y_{JI_J}, x_{J1} \ldots x_{JI_J})} \quad (7)$$

$$= \frac{P(y_{11} \ldots y_{1I_1}, \ldots, y_{J1} \ldots y_{JI_J}|x_{11} \ldots x_{1I_1}, \ldots, x_{J1} \ldots x_{JI_J}, \theta_1 \ldots \theta_J, \theta^*)P(x_{11} \ldots x_{1I_1}, \ldots, x_{J1} \ldots x_{JI_J}, \theta_1 \ldots \theta_J, \theta^*)}{P(y_{11} \ldots y_{1I_1}, x_{11} \ldots x_{1I_1}, \ldots, y_{J1} \ldots y_{JI_J}, x_{J1} \ldots x_{JI_J})} \quad (8)$$

$$= \frac{P(y_{11} \ldots y_{1I_1}, \ldots, y_{J1} \ldots y_{JI_J}|x_{11} \ldots x_{1I_1}, \ldots, x_{J1} \ldots x_{JI_J}, \theta_1 \ldots \theta_J, \theta^*)P(x_{11} \ldots x_{1I_1}, \ldots, x_{J1} \ldots x_{JI_J})P(\theta_1 \ldots \theta_J, \theta^*)}{P(y_{11} \ldots y_{1I_1}, x_{11} \ldots x_{1I_1}, \ldots, y_{J1} \ldots y_{JI_J}, x_{J1} \ldots x_{JI_J})} \quad (9)$$

$$= \frac{\left(\prod_{j=1}^{J} P(y_{j1} \ldots y_{jI_j}|x_{j1} \ldots x_{jI_j}, \theta_j)P(\theta_j|\theta^*)\right)P(x_{11} \ldots x_{1I_1}, \ldots, x_{J1} \ldots x_{JI_J})P(\theta^*)}{P(y_{11} \ldots y_{1I_1}, x_{11} \ldots x_{1I_1}, \ldots, y_{J1} \ldots y_{JI_J}, x_{J1} \ldots x_{JI_J})} \quad (10)$$

as $P(y_{kji}|x_{ji}, \theta_j)$ for readability. We now describe how maximum a posteriori estimation is used to find the parameters $\theta_1, \ldots, \theta_J$ corresponding to each source domain and $\theta_*$ which is a center model that ties all the source models together. Eqs. 6-8. give the probability of the parameters given the data. In eq. 9. we separate the dependence on the input and the model. In eq. 10. we separate the dependence between each domain, where the weights $\theta_j$ are relevant for domain $D_j$, and each model $\theta_j$ has a prior with mean $\theta^*$. Maximum a posterior estimation tries to maximize the posterior distribution given in eqs. 11. and 12. below. Notice again, eq. 12. is reflected in the Bayesian network in Fig. 1.

As noted above in eq. 4., each prediction is a categorical distribution with class probability parameters given by a model $\hat{y}_{kji} = F_k(x_{ji}, \theta_j)$ for domain j, sample i, class k. Note that the categorical probabilities must sum to 1, ($\sum_{k=1}^{K} \hat{y}_{kji} = 1$). Therefore, the final layer of each model $F(x_{ji}, \theta_j)$ is a softmax layer representing a class distribution prediction. The prior distributions of the model parameters $\theta_j$ are multivariate normal ($\theta_j$ is a vector with $N$ parameters) with mean $\theta_*$ and precision

$$\underset{\theta_1,\ldots,\theta_J,\theta_*}{\text{argmax}} \frac{\left(\prod_{j=1}^{J} P(y_{j1} \ldots y_{jI_j}|x_{j1} \ldots x_{jI_j},\theta_j)P(\theta_j|\theta^*)\right)P(x_{11}\ldots x_{1I_1},\ldots,x_{J1}\ldots x_{JI_J})\,P(\theta^*)}{P(y_{11}\ldots y_{1I_1},x_{11}\ldots x_{1I_1},\ldots,y_{J1}\ldots y_{JI_J},x_{J1}\ldots x_{JI_J})} \quad (11)$$

$$= \underset{\theta_1,\ldots,\theta_J,\theta_*}{\text{argmax}} \left\{\left(\prod_{j=1}^{J} P(y_{j1}\ldots y_{jI_j}|x_{j1}\ldots x_{jI_j},\theta_j)P(\theta_j|\theta^*)\right)P(\theta^*)\right\} \quad (12)$$

$$= \underset{\theta_1\ldots\theta_J,\theta^*}{\text{argmax}} \left\{\left(\prod_{j=1}^{J}\left(\prod_{i=1}^{I_j} cat(y_{ji}|F(x_{ji},\theta_j))\right)MVN(\theta_j|\theta^*,(2\lambda_1)^{-1}I)\right)MVN(\theta^*|0,(2\lambda_2)^{-1}I)\right\} \quad (13)$$

fixed a priori as $2\lambda_1$. The prior distribution of $\theta_*$ is multivariate normal with mean 0 and precision $2\lambda_2$, which is equivalent to weight regularization. This is shown in eq. 13. where each probability in eq. 12. is shown as its respective distribution function. In appendix A, we continue from eq. 13. and derive the Hierarchical Bayesian MDA model loss function which is shown to be a regularized cross entropy. This loss function is given from the appendix as eq. 21 below. After using the loss function to train the models $F(x_j,\theta_j)$, we can use these models to predict labels for each source domain $D_j$, similar to the multitask model in section IIA, such that $F_j: \mathcal{X} \to \mathcal{L}$.

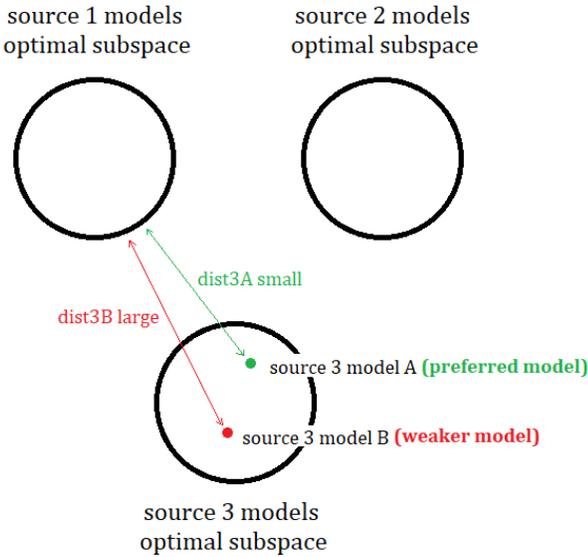

Fig. 2. Optimal Source Model Subspaces

In Fig. 2. we visualize the subspaces for the optimal source models for each source domain. If a model is randomly initialized and trained for a source domain dataset several times, there will be a subspace of models that have optimal performance, represented by the circles. The Hierarchical Bayesian model prefers models that are more similar. For example, if we have source 3 model A and source 3 model B as shown in the figure, both models will be optimal for source 3 data. However, because model A is closer the subspace of optimal source 1 models ($dist3A < dist3B$ in Fig. 2.) and to the subspace of optimal source 2 models, we expect model A to outperform model B on these other source domains. Using the Bayesian loss function, we are penalizing differences between $\theta_j$ and $\theta_*$ for all source domains $j$. This means we are selecting models for each source domain that are expected to perform well on the other domains. Finally, because models $\theta_1,\ldots,\theta_J$ perform better across domains, we expect better target performance than if we selected models that were only optimized to their respective source domain.

In Fig. 3. we highlight the difference in this approach vs. existing methods in the literature. Some methods use a single source encoder with shared weights for all source domains, represented on the left on the spectrum. Other methods use independent source encoders for each source domain, represented on the right of the spectrum. The Bayesian method searches for the optimal balance by varying the hyperparameter $\lambda_1$ from the loss in eq. 13. The higher the precision or higher $\lambda_1$, the more we penalize models that are different, meaning we are closer to the left side of the spectrum. The lower the precision, or lower $\lambda_1$, the less we penalize models that are different, meaning we are closer to the right side of the spectrum. By varying $\lambda_1$ we are able to empirically find a balance, which generalizes the two choices of shared or independently trained weights to a middle ground we select for by measuring testing target accuracy.

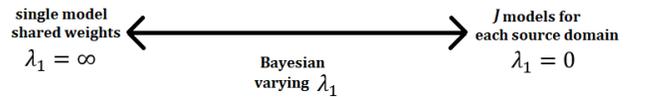

Fig. 3. Bayesian Generalization over Single Model vs $J$ Models

### B. Framework Applied to Encoder Decoder Models

We can apply our framework to training of encoders and decoders for models with a single encoder [9] [12] [14] [16] or $J$ encoders for each source domain [10] [11] [15]. The optimal balance as described above is obtained by varying the balance using $\lambda_1$ hyperparameter. Using the loss function given earlier in eq. 21., we show explicitly in eq. 22. how to use this loss function from our Bayesian framework when the encoders and decoders need to be isolated as in [9] [10] [11] [12] [14] [15] [16] [18]. See also eq. 4. in section IIIA. where the models $F_j$

$$loss(\theta_1\ldots\theta_J,\theta_*,X_1,Y_1,\ldots,X_J,Y_J) = \sum_{j=1}^{J}\sum_{i=1}^{I_j}\sum_{k=1}^{K} y_{kji}\ln\left(\frac{1}{F_k(x_{ji},\theta_j)}\right) + \lambda_1\sum_{j=1}^{J}\sum_{n=1}^{N}(\theta_{jn}-\theta_{*n})^2 + \lambda_2\sum_{n=1}^{N}\theta_{*n}^2 \quad (21)$$

that we are optimizing using the loss in eq. 21. are defined in terms of the operation or composition of encoders $F_{Ej}$ and decoders $F_{Cj}$.

$$loss(\theta_{E1} \ldots \theta_{EJ}, \theta_{E*}, \theta_{C1} \ldots \theta_{CJ}, \theta_{C*}, X_1, Y_1, \ldots, X_J, Y_J) =$$

$$\sum_{j=1}^{J} \sum_{i=1}^{I_j} \sum_{k=1}^{K} y_{kji} \ln\left(\frac{1}{F_{C_k}(F_E(x_{ji}|\theta_{Ej}),\theta_{Cj})}\right) +$$

$$\lambda_1 \sum_{j=1}^{J} \sum_{n=1}^{N_E} (\theta_{Ejn} - \theta_{E*n})^2 + \lambda_2 \sum_{n=1}^{N_E} \theta_{E*n}^2 +$$

$$\lambda_1 \sum_{j=1}^{J} \sum_{n=1}^{N_C} (\theta_{Cjn} - \theta_{C*n})^2 + \lambda_2 \sum_{n=1}^{N_C} \theta_{C*n}^2 \qquad (22)$$

### C. Confidence Based Target Domain Inference

The Hierarchical Bayesian Framework is optimized using all source domain datasets. There are various approaches that can be used to perform target domain testing given the $J$ source models that are obtained. In Deep Ensemble Learning [10] the source classifier with the maximum confidence for each target sample is used to predict the class label. In Deep Cocktail Network [9] there are $J$ discriminators between the target and every source domain. The discriminator confidences are used to weigh the classifier outputs in a linear combination that predicts the label for each target sample. Similarly, in our method we use each source model $\theta_j$ to predict a probability distribution $\hat{y}_{kji}$ over the labels $l_k$ given a target sample $x_{ti}$. The categorical distribution with the most confidence, meaning the lowest entropy is selected and used to predict the target label. The reasoning is that when a target sample is most similar to the samples from source domain $D_j$, or if classifier $F_j$ has the most discriminative power and therefore lowest entropy, then $F_j$ is the most likely source classifier to make the correct label prediction.

### D. Integration with Cross-Attention Action Recognition

To test our Bayesian framework, we work on solving the challenging MDA problem of human action recognition in RGB video. We utilize a state-of-the-art video action recognition model, Cross-Attention Action Recognition (CAST) [29]. This model uses a two-stream architecture that trains a spatial attention expert and a temporal attention expert. The two parallel streams exchange information before their final layer which allows spatial and temporal features to influence each other.

When using the CAST model for each source domain, we initialize with pretrained weights for Kinetics-400 human action recognition dataset [30]. These weights are provided by the designers of CAST [29] on their GitHub project page [31]. For now, we drop the weight norm regularization of the center model, which is the term governed by the hyperparameter $\lambda_2$ from eq. 13 and eq. 21. above. This reduces the grid search size for the hyperparameter search and also avoids reducing the weight norms below the region that was found when the CAST model was originally tuned. The CAST model uses AdamW optimizer from the PyTorch library. The Bayesian framework required us to add a second optimizer for the $\lambda_1$ coefficient loss terms between the center $\theta^*$ and source models $\theta_j$. We used stochastic gradient descent (SGD), again using PyTorch.

### E. Hierarchical Bayesian vs Limiting Cases

To demonstrate the effectiveness of the Hierarchical Bayesian Framework, we perform a hyperparameter search, where we vary $\lambda_1$ as mentioned in the previous sections IIIA and IIID. This allows us to perform an ablation study, but also prove the effectiveness of the Hierarchical Bayesian Framework. As we increase $\lambda_1$ we approach a single model with shared weights for all source domains as in [9] [12] [14] [16]. As we reduce $\lambda_1$ to 0, we are back to $J$ independently trained models for each source domain $J$ [10] [11] [15]. Note that for each of the methods from the literature [9] [10] [11] [12] [14] [15] [16], our framework is shown to obtain identical results, since the these other methods may be considered as edge case of our proposed framework with either single encoder or independent encoders. The single model first edge case is obtained by actually training a single model with shared weights to show the performance of that method. The independent model edge case is obtained by training $J$ models independently to show the performance of that method as well.

### F. Daily-DA Benchmark Dataset

The Daily-DA RGB video dataset has been used as a benchmark for human action recognition tasks, and was introduced in the paper on Temporal Attentive Moment Alignment Network [12] to show performance on the MDA task. It was later used in Source-Free Domain Adaptation [13] to solve a related problem of UDA where the source data is not available (source-free) when the unlabeled target data is presented. The data can be obtained by downloading each domain separately as described at the Daily-DA project page (https://xuyu0010.github.io/msvda.html). This data contains different classes per each subdomain: ARID, HMDB51, Moments in Time, and Kinetics-600. The Daily-DA benchmark limits the action classes to 8 actions that are present in the intersection of actions for all of the subdomains. These classes are drink, jump, pick, pour, push, run, walk, and wave. In the original Daily-DA benchmark paper, the class names for each subdomain are given in their Table II [12]. Alternatively, we have provided the preprocessed data that contains these 8 classes. We have normalized all the data already to obtain a tensor of Nx224x224x3 for each video sample of N frames, height and width of 224 and 3 channels for RGB. This prepared dataset can be obtained upon request by filling out the form at the ODU Vision Lab website external resources page with the following link (https://sites.wp.odu.edu/VisionLab/resources).

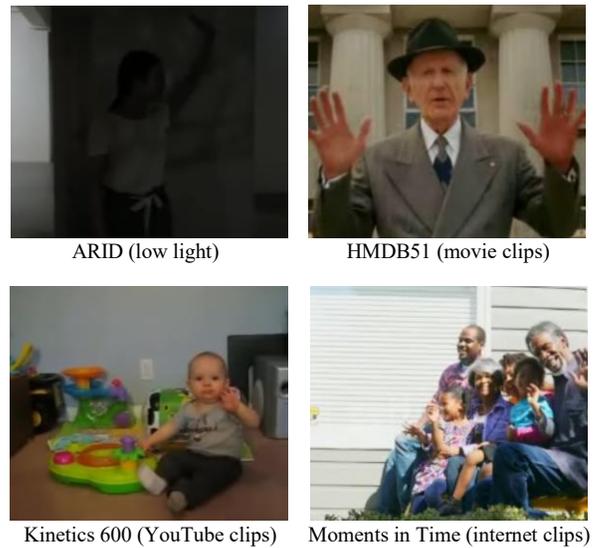

| ARID (low light) | HMDB51 (movie clips) |
| Kinetics 600 (YouTube clips) | Moments in Time (internet clips) |

Fig. 4. Daily-DA benchmark human action recognition dataset

## IV. RESULTS AND DISCUSSION

Here we describe our experiment on human action recognition in RGB videos using our Hierarchical Bayesian Framework. We report performance using accuracy on the multiclass target domain classification. Our results are compared to the existing benchmark performance on the Daily-DA data.

### A. Experimental Design

To train the Hierarchical Bayesian model using the CAST architecture for all target domain problems requires over 6 days on a high-performance cluster using an Nvidia V100 GPU with 128GB RAM and 16GB VRAM for the GPU data. To perform our ablation study, we needed to repeat this training for each hyperparameter setting. Finally, after finding the best hyperparameter setting, we increased the number of epochs in the pretraining of the source models which required another 12 days. Table I gives the number of videos in each domain in the Daily-DA dataset. These domains are imbalanced, but all 4 domains have a large number of sample videos. Our preprocessing of selecting the 8 classes assigned gives similar subdomain sizes to [12].

TABLE I. DAILY-DA SUBDOMAINS

| Subdomain | Number of Video Samples |
|---|---|
| ARID (A) | 4552 |
| HMDB51 (H) | 1527 |
| Moments in Time (M) | 27248 |
| Kinetics-600 (K) | 9329 |

### B. Multisource Domain Adaptation Results

Table II shows our Hierarchical Bayesian model performance using target domain classification accuracy. We report the accuracy for each subdomain separately and report the average in the right column. For example, column A means the source domains are HMD51, Moment in Time, and Kinetics-600 and that the target domain is ARID. Note that the center model is able to predict the target domain class labels. As shown in Fig. 2., since the center model is trained to perform on all source domains, the generalization to the target domain is reasonable. These results are reported at the top of the table. The improved method using confidence-based prediction selection as described in section IIIC is reported below the center-based predictions in the table. (There is no center prediction given for the shared weight model since, that is the same as using the confidence in that case.) In the ablation study we trained for 70 epochs to search for the hyperparameter value for $\lambda_1$ that maximizes test performance on the target domain. Finally, using the best hyperparameter value of $\lambda_1 = 1 \cdot 10^{-3}$ we continue training to a total of 140 epochs. We do not see addition performance gains with the addition training. This is reported in Table II.

TABLE II. HIERARCHICAL BAYESIAN PERFORMANCE

| Ablation with 70 Epochs (center model prediction) | Accuracy for each Target Domain | | | | |
|---|---|---|---|---|---|
| | A | H | M | K | Average |
| $\lambda_1 = \infty$ (single model) | N/A | N/A | N/A | N/A | N/A |
| $\lambda_1 = 1 \cdot 10^{-1}$ | 42.29 | 67.10 | 36.42 | 70.88 | 54.17 |
| $\lambda_1 = 1 \cdot 10^{-3}$ | 28.19 | 64.47 | 14.99 | 25.27 | 33.23 |
| $\lambda_1 = 1 \cdot 10^{-5}$ | 14.10 | 7.90 | 12.41 | 14.99 | 12.35 |
| $\lambda_1 = 0$ (independent models) | 10.57 | 10.53 | 12.41 | 8.14 | 10.41 |
| Ablation with 70 Epochs (confidence-based prediction) | Accuracy for each Target Domain | | | | |
| | A | H | M | K | Average |
| $\lambda_1 = \infty$ (single model) | 56.27 | 76.84 | 15.00 | 47.01 | 48.78 |
| $\lambda_1 = 1 \cdot 10^{-1}$ | 53.49 | 68.83 | 47.54 | 70.28 | 60.04 |
| $\lambda_1 = 1 \cdot 10^{-3}$ | 54.65 | 71.96 | 53.48 | 68.45 | **62.14** |
| $\lambda_1 = 1 \cdot 10^{-5}$ | 54.54 | 72.24 | 15.22 | 17.09 | 39.77 |
| $\lambda_1 = 0$ (independent models) | 56.01 | 70.92 | 15.81 | 16.45 | 39.80 |
| Final Test with 140 Epochs | Accuracy for Optimal Hyperparameter | | | | |
| | A | H | M | K | Average |
| $\lambda_1 = 1 \cdot 10^{-3}$ | 53.00 | 71.27 | 15.74 | 29.48 | 42.37 |
| Previous State-of-the-Art Method | Benchmark Result from Literature | | | | |
| | A | H | M | K | Average |
| Attention and Moment Alignment [12] | 29.95 | 48.33 | 36.75 | 64.36 | 44.85 |

### C. Comparison with Existing MDA Methods

The Hierarchical Bayesian Framework performs a similar task to existing works that attempt domain alignment using addition terms in the loss (for example feature moment alignment, feature difference or class difference penalties, or domain discriminators) [9] [10] [11] [12] [14] [15] [16] [18]. Our Bayesian Framework performs feature alignment as the encoder weights of each source model $F_j$ are optimized to reduce the distance between models $\theta_{Ej_1}$ and $\theta_{Ej_2}$ for different source domains. These are the loss terms with the $\lambda_1$ coefficient in eq. 22. Our Bayesian Framework also performs class alignment as the end-to-end loss for each source model pair $F_{j_1}$, $F_{j_2}$ is minimized when the predicted class distribution is correct for both source domains $D_{j_1}$ and $D_{j_2}$. The added benefit of our Bayesian Framework over the existing methods is that we find an initial set of models $F_{j_1}$ trained on each source domain $D_{j_1}$ that are also optimized to perform on the remaining source domains $D_{j_2 \neq j_1}$. This allows us to achieve much higher performance than the existing work on the Daily-DA benchmark. The best accuracy on MDA action recognition using video data benchmark Daily-DA is 44.85% accuracy [12] and our model improves the accuracy to 62.14%.

Further note that for each model setup in the literature our method will either outperform for a specific value of the $\lambda_1$ hyperparameter or in the worst case our model will at least match the existing result with $\lambda_1 = \infty$ for a single model or $\lambda_1 = 0$ for independent models.

## V. CONCLUSION

We conclude by reviewing the contributions of this work as follows. First, we analyzed the existing methods for multisource domain adaptation (MDA) and found that these methods either use a single source model [9] [12] [14] [16] or $J$ independent source models for $J$ source domains [10] [11] [15]. We then developed a Bayesian Framework to allow us to consider a more general scenario with $J$ dependent source models using a Hierarchical Bayesian model. This model is derived using a Bayesian network that models the problem distribution and is factored into a maximum a posteriori estimation problem. Finally, this estimation problem is solved with a loss function that optimizes the parameters of $J$ Hierarchically trained source models given the source data. The loss function takes the form of cross entropy with additional regularization terms to encourage the source model for each domain to perform well on all other source domains. We show that the Hierarchical Bayesian Framework outperforms naïve single model and naïve $J$ independent source models on the challenging problem of human action recognition in RGB videos for MDA. We achieve 62.14% accuracy on the benchmark Daily-DA data which is a 17.29% improvement over existing works in the literature. Further note that existing results [9] [10] [11] [12] [14] [15] [16] can be considered? as edge cases or improved using our Hierarchical Bayesian Framework. We emphasize that this means that our Bayesian Framework can be combined with other deep network approaches to improve existing methods for other target domains. It is also possible to consider multiple target domains [6] [7], but this is outside the current scope of our problem. In Multi-Modal UDA [8], the source domains are from different sensor modalities. This can be addressed with our proposed method, or more generally by using models with different architectures for each modality.

For future work, it would be interesting to test the framework on other MDA datasets, for example Sport-DA. We have saved this for future work, as training each hyperparameter setting takes over 6 days on Nvidia V100 GPU, adding up to a month of training time to perform the ablation study and final fine-tuning. We also would like to perform an addition training step with a loss term using target input samples to align source and target domains. This could further improve the MDA action recognition accuracy.

Appendix A

Here we take the maximum a posterior parameter estimation equation from eq. 13. in section IIIA above and derive the Hierarchical Bayesian loss function for multisource domain adaptation. Eqs. 14-16 use natural log to simplify the loss function.

$$\underset{\theta_1 \ldots \theta_J, \theta_*}{\operatorname{argmax}} \left\{ \left( \prod_{j=1}^{J} \left( \prod_{i=1}^{I_j} cat\left(y_{ji} \middle| \mathrm{F}(x_{ji}, \theta_j)\right) \right) MVN(\theta_j | \theta_*, (2\lambda_1)^{-1}\mathrm{I}) \right) MVN(\theta^* | 0, (2\lambda_2)^{-1}\mathrm{I}) \right\} \quad (13)$$

$$= \underset{\theta_1 \ldots \theta_J, \theta_*}{\operatorname{argmax}} \left\{ \ln\left( \left( \prod_{j=1}^{J} \left( \prod_{i=1}^{I_j} cat\left(y_{ji} \middle| \mathrm{F}(x_{ji}, \theta_j)\right) \right) MVN(\theta_j | \theta_*, (2\lambda_1)^{-1}\mathrm{I}) \right) MVN(\theta_* | 0, (2\lambda_2)^{-1}\mathrm{I}) \right) \right\} \quad (14)$$

$$= \underset{\theta_1 \ldots \theta_J, \theta_*}{\operatorname{argmax}} \left\{ \ln\left( \left( \prod_{j=1}^{J} \left( \prod_{i=1}^{I_j} \prod_{k=1}^{K} \left( \mathrm{F}_k(x_{ji}, \theta_j) \right)^{y_{kji}} \left(\frac{1}{\pi}\right)^{\frac{N}{2}} \left(\frac{1}{|2\lambda_1 \mathrm{I}|}\right)^{\frac{1}{2}} \exp\left(-\lambda_1 (\theta_j - \theta_*)^T (\theta_j - \theta_*)\right) \right) \left(\frac{1}{\pi}\right)^{\frac{N}{2}} \left(\frac{1}{|2\lambda_2 \mathrm{I}|}\right)^{\frac{1}{2}} \exp(-\lambda_2 \theta_*^T \theta_*) \right) \right\} \quad (15)$$

$$= \underset{\theta_1 \ldots \theta_J, \theta_*}{\operatorname{argmax}} \left\{ \left( \sum_{j=1}^{J} \left( \sum_{i=1}^{I_j} \sum_{k=1}^{K} y_{kji} \ln\left( \mathrm{F}_k(x_{ji}, \theta_j) \right) \right) + \frac{N}{2} \ln\left(\frac{1}{\pi}\right) + \frac{1}{2} \ln\left(\frac{1}{|2\lambda_1 \mathrm{I}|}\right) - \lambda_1 (\theta_j - \theta_*)^T (\theta_j - \theta_*) \right) + \frac{N}{2} \ln\left(\frac{1}{\pi}\right) + \frac{1}{2} \ln\left(\frac{1}{|2\lambda_2 \mathrm{I}|}\right) - \lambda_2 \theta_*^T \theta_* \right\} \quad (16)$$

Next, considering only terms that depend on $\theta_1 \ldots \theta_J, \theta_*$ we simplify further in eqs. 16-20. Where $n \in \{1, \ldots, N\}$ are the weight indices of the parameters $\theta_1 \ldots \theta_J, \theta_*$, since each weight vector for the hierarchical model has the same dimension.

Therefore, the loss function for optimizing our models $F(x_j, \theta_j)$ and $\theta_*$ is given in eq. 21. This is regularized, Hierarchical Bayesian loss for multisource domain adaptation with $J$ source domains. The first term is cross entropy between source sample targets $y_{kji}$ and model predictions $F_k(x_{ji}, \theta_j)$. The second term encourages a model $\theta_*$ to have weights that are similar to each model $\theta_j$. The third term regularizes the weights of the model $\theta_*$ directly and also regularizes the weights of each source model $\theta_j$ indirectly.

$$loss(\theta_1 \ldots \theta_J, \theta_*, X_1, Y_1, \ldots, X_J, Y_J) = \sum_{j=1}^{J} \sum_{i=1}^{I_j} \sum_{k=1}^{K} y_{kji} \ln\left(\frac{1}{F_k(x_{ji},\theta_j)}\right) + \lambda_1 \sum_{j=1}^{J} \sum_{n=1}^{N} (\theta_{jn} - \theta_{*n})^2 + \lambda_2 \sum_{n=1}^{N} \theta_{*n}^2 \quad (21)$$

$$\underset{\theta_1 \ldots \theta_J, \theta_*}{\operatorname{argmax}} \left\{ \left( \sum_{j=1}^{J} \left( \sum_{i=1}^{I_j} \sum_{k=1}^{K} y_{kji} \ln\left(F_k(x_{ji}, \theta_j)\right)\right) + \frac{N}{2} \ln\left(\frac{1}{\pi}\right) + \frac{1}{2} \ln\left(\frac{1}{|2\lambda_2 I|}\right) - \lambda_1 (\theta_j - \theta_*)^T (\theta_j - \theta_*)\right) + \frac{N}{2} \ln\left(\frac{1}{\pi}\right) + \frac{1}{2} \ln\left(\frac{1}{|2\lambda_2 I|}\right) - \lambda_2 \theta_*^T \theta_* \right\} \quad (16)$$

$$= \underset{\theta_1 \ldots \theta_J, \theta_*}{\operatorname{argmax}} \left\{ \left( \sum_{j=1}^{J} \left( \sum_{i=1}^{I_j} \sum_{k=1}^{K} y_{kji} \ln\left(F_k(x_{ji}, \theta_j)\right)\right) - \lambda_1 (\theta_j - \theta_*)^T (\theta_j - \theta_*)\right) - \lambda_2 \theta_*^T \theta_* \right\} \quad (17)$$

$$= \underset{\theta_1 \ldots \theta_J, \theta_*}{\operatorname{argmin}} \left\{ \left( \sum_{j=1}^{J} \left( \sum_{i=1}^{I_j} \sum_{k=1}^{K} y_{kji} \ln\left(\frac{1}{F_k(x_{ji},\theta_j)}\right)\right) + \lambda_1 (\theta_j - \theta_*)^T (\theta_j - \theta_*)\right) + \lambda_2 \theta_*^T \theta_* \right\} \quad (18)$$

$$= \underset{\theta_1 \ldots \theta_J, \theta_*}{\operatorname{argmin}} \left\{ \left( \sum_{j=1}^{J} \left( \sum_{i=1}^{I_j} \sum_{k=1}^{K} y_{kji} \ln\left(\frac{1}{F_k(x_{ji},\theta_j)}\right)\right) + \lambda_1 \sum_{n=1}^{N} (\theta_{jn} - \theta_{*n})^2 \right) + \lambda_2 \sum_{n=1}^{N} \theta_{*n}^2 \right\} \quad (19)$$

$$= \underset{\theta_1 \ldots \theta_J, \theta_*}{\operatorname{argmin}} \left\{ \sum_{j=1}^{J} \sum_{i=1}^{I_j} \sum_{k=1}^{K} y_{kji} \ln\left(\frac{1}{F_k(x_{ji},\theta_j)}\right) + \lambda_1 \sum_{j=1}^{J} \sum_{n=1}^{N} (\theta_{jn} - \theta_{*n})^2 + \lambda_2 \sum_{n=1}^{N} \theta_{*n}^2 \right\} \quad (20)$$